\documentclass[
]{ceurart}

\usepackage{graphicx}

\usepackage{listings}
\usepackage{nicefrac}
\usepackage{xcolor}

\usepackage{ulem}

\begin{document}

\copyrightyear{2025}
\copyrightclause{Copyright for this paper by its authors.
  Use permitted under Creative Commons License Attribution 4.0
  International (CC BY 4.0).}

\conference{AIC2025: 10th Workshop on Artificial Intelligence and Cognition, October 25--26, 2025, jointly with ECAI 2025, Bologna, Italy
}

\title{Exploring structures of inferential mechanisms\\through simplistic digital circuits}


\author[1]{Giovanni Sileno}[%
orcid=0000-0001-5155-9021,
email=g.sileno@uva.nl,
url=https://gsileno.net/,
]
\cormark[1]
\address[1]{Informatics Institute, University of Amsterdam}

\author[2,3]{Jean-Louis Dessalles}[%
orcid=0000-0002-3910-4611,
email=dessalles@telecom-paris.fr,
url=https://perso.telecom-paristech.fr/jld/,
]
\address[2]{Télécom Paris, Paris, France}
\address[3]{Institut Polytechnique de Paris, Paris, France}


\cortext[1]{Corresponding author.}

\begin{abstract}
Cognitive studies and artificial intelligence have developed distinct models for various inferential mechanisms (categorization, induction, abduction, causal inference, contrast, merge, ...). Yet, both natural and artificial views on cognition lack apparently a unifying framework. This paper formulates a speculative answer attempting to respond to this gap. To postulate on higher-level activation processes from a material perspective, we consider inferential mechanisms informed by symbolic AI modelling techniques, through the simplistic lenses of electronic circuits based on logic gates. We observe that a logic gate view entails a different treatment of implication and negation compared to standard logic and logic programming. Then, by combinatorial exploration, we identify four main forms of dependencies that can be realized by these inferential circuits. Looking at how these forms are generally used in the context of logic programs, we identify eight common inferential patterns, exposing traditionally distinct inferential mechanisms in an unifying framework. Finally, following a probabilistic interpretation of logic programs, we unveil inner functional dependencies. The paper concludes elaborating in what sense, even if our arguments are mostly informed by symbolic means and digital systems infrastructures, our observations may pinpoint to more generally applicable structures.
\end{abstract}


\maketitle

\section{Introduction}

Cognitive studies have distinguished several types of cognitive mechanisms, by conducting distinct 
modelling efforts and different types of experiments. For instance, categorization (the process by which humans group objects, events, situations, on the basis of shared characteristics) has been approached through rule-based models \cite{Bruner1966}, prototype theory \cite{Rosch1978}, exemplar theory \cite{Nosofsky1986}, knowledge-based models \cite{MurphyMedin1985}, and with Bayesian inference \cite{Goodman2008}. Induction, the process by which we humans draw general rules from observations, has been approached via associative models based on co-occurrence \cite{Allan1993}, descriptive models based on similarity \cite{Sloman1993}, and again through Bayesian models \cite{PerforsTenenbaumGriffiths2006}. 
Similar diversification appears for other cognitive processes like abduction (the process of inferring the best explanation for a set of observations) \cite{Lipton2004}, deductive reasoning \cite{Rips1994}, analogical reasoning \cite{Gentner1983}, causal/diagnostic inference \cite{WaldmannHolyoak1992}, conceptual contrast \cite{MarkmanGentner1993,Dessalles2015}, conceptual merge/blending \cite{FauconnierTurner1998}, and so on. Complementary to these streams of works, AI research and practice have been developing for decades distinct symbolic and sub-symbolic methods aiming to reproduce these cognitive functions by computational means. For instance, in symbolic AI, categorization has been approached relying on rules in rule-based systems \cite{BuchananShortliffe1984}, on (methods constructing) decision trees \cite{Quinlan1986}, and on formal concept analysis \cite{Wille1982}; induction has been approached via inductive logic programming \cite{MuggletonDeRaedt1994}, version-space \cite{Mitchell1982} and explanation-based learning \cite{MitchellKellerKedar-Cabelli1986}. In sub-symbolic AI, categorization has been approached by means of neural networks \cite{Bishop2006}, support vector machines \cite{CristianiniShawe-Taylor2000}, as well as through clustering algorithms \cite{AggarwalReddy2014}. Intermediate proposals also exist, for instance based on conceptual spaces \cite{Gardenfors2001}, aiming to provide both intelligibility and perceptual grounding. Inducing structure from data has been formally connected in algorithmic information theory \cite{Solomonoff1964} to the \textit{minimum description length} principle, formalizing Occam's razor. 
In practice, induction is at the core of all machine learning methods, including deep learning and generative AI methods. With contemporary transformer architectures \cite{Vaswani2017}, there seems to be a general belief that, by getting induction right, the inferential mechanisms appropriate to solving the task (or any task) will be induced by training, given an adequate amount of data. This assumption explains the renewed interest in reverse engineering neural networks to study activation patters, as e.g. in \textit{mechanicistic interpretation} \cite{Bereska2024}. 

With hindsight, diversification in both disciplines has been crucial to improve prediction accuracy (on the modelling side), as well as efficiency and effectiveness of performance (on the design side). Yet, such a functional diversification has not promoted the identification of a simple unifying framework that explains where these functions emerge from. In this paper, we will approach this question from a systematic perspective, focusing on a simplistic artificial system relying on logic gates (an ideal digital electronic circuit) meant to perform inferences (section 2). Building upon this basis, an unexpectedly unifying framework is obtained (section 3) by performing a combinatorial exploration of the possible activation dependencies, informed by common modelling practices in logic and symbolic AI. By discussing the assumptions and constraints of this speculative exercise, we conclude that our elaboration may pinpoint to more generally applicable structures.
An extension integrating axioms from probability theory is then discussed (section 4), exposing dependencies across dependencies. 

\section{Going electronical}

Methods attempting to reverse engineer the inferential constructs expressed by large neural models are gaining traction in the literature  \cite{Bereska2024}. These works share a perspective similar {to} empirical neuroscience, trying to get the functioning of the mind by studying its physical form, or its implementation level. 
The present paper starts instead from a complementary perspective. Despite known scaling limitations and arguable cognitive assumptions, several inferential mechanisms have been reproduced with success through symbolic AI methods. This could suggest that, if these methods perform at least in part as humans would do, they get some aspect of the cognitive functions right, at least in those contexts. Note that the \textit{Physical Symbol System} and \textit{Language of Thought} (LoT) hypotheses go beyond this functional alignment --- but we do not need to abide by stronger assumptions. The motivation behind the present work comes from the intuition that grounding inferential mechanisms on simplistic hardware processes may help reduce assumptions that come along with symbolic processing.

\subsection{Logic gate circuits}

Logic gate circuits are at the basis of digital electronics. By composing primitive electronic components reproducing logical functions (AND, OR, XOR, ...), they perform operations on Boolean inputs, resulting in Boolean outputs. From this perspective, they can be seen as materially realizing logical inference. Yet, strictly speaking, there are no symbols involved at the hardware level: tensions and currents are all what is being processed. Rather, logic gate circuits can be seen as a specific instances of a network, whose nodes are \textit{activated} depending on the state of other nodes and the type of connections binding them. For this reason, they can also be seen as a very simplistic version of a neural network, although without weights nor continuous activation functions. 

\hypertarget{logic-programs}{%
\subsection{Logic programs}\label{logic-programs}}

\begin{figure}[t]
\centering
\scalebox{.27}{
\includegraphics{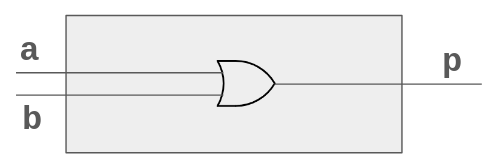}
}
\caption{A simple digital circuit (a single AND port).}
\label{fig:simplecircuit}
\end{figure}

Let us consider the simple digital circuit (a single AND port) in Fig.~\ref{fig:simplecircuit}.
One could easily associate the function implemented by this circuit to the following logic rule (written in Prolog/ASP syntax):
\begin{lstlisting}
p :- a, b.
\end{lstlisting}
which in turn corresponds to the logical dependency:
$$ a \wedge b \rightarrow p $$
In standard logic, however, the conditional above holds together with its contrapositive:
$$ \neg p \rightarrow \neg a \vee \neg b $$
The conditional expressed in the contrapositive is however not directly implementable in a digital
circuit: the output is not a single port, and it is non-deterministic.
To construct a valid circuit, we need to remove the source of
non-determinism, obtaining the following formulas (in ASP, with the \textit{strong negation} operator ``\texttt{-}''):
\begin{lstlisting}
p :- a, b.
-b :- -p, a.
-a :- -p, b.
\end{lstlisting}

Interestingly, the fastest derivation of these three rules comes from interpreting the conditional as \textit{material implication}:
$$ a, b \rightarrow p \quad \Leftrightarrow \quad \neg (a \wedge b) \vee p \leftrightarrow  \neg ( a \wedge b \wedge \neg p ) $$
which would be expressed in ASP as the constraint:
\begin{lstlisting}
:- a, b, -p.
\end{lstlisting}
meaning that \(a\) and \(b\) and \(\neg p\) cannot be simultaneously
true. Thus, the conditional illustrated by the first circuit emerges from the
constraint derived from material implication as the only circuit whose
components are in the ``right'' place (positive atoms in the body of the
rule, negated atom in the head).\footnote{The constraint corresponds to the negation of a \textit{Horn clause} --- logic formulas that contain at most a positive literal --- known to have useful properties for automated reasoning, and for this reason at the base of logic programming approaches.}

\hypertarget{logical-systems}{%
\subsection{Logical systems}\label{logical-systems}}

Given any propositional variable, logic deals with relations concerning
both the true and the false state of this variable. This interpretation becomes
manifest when we introduce an arbitrary conditional, for the simultaneous holding of its contrapositive. To force the realization of these constraints in
all possible configurations of the world, we need to add to the deterministic machinery above
a combinatorial exploration of all possible states, as expressed in the
following ASP program:
\begin{lstlisting}
p :- a, b.
-a :- -p, b.
-b :- -p, a.
1{a; -a}1. 1{b; -b}1. 1{p; -p}1.
\end{lstlisting}
The answer set provided by the solver (e.g. \texttt{clingo}) corresponds to the set of logically possible models, consequently to the constraint, concerning the three variables and their negations.

\begin{figure}[t]
\centering
\scalebox{0.27}{
\includegraphics{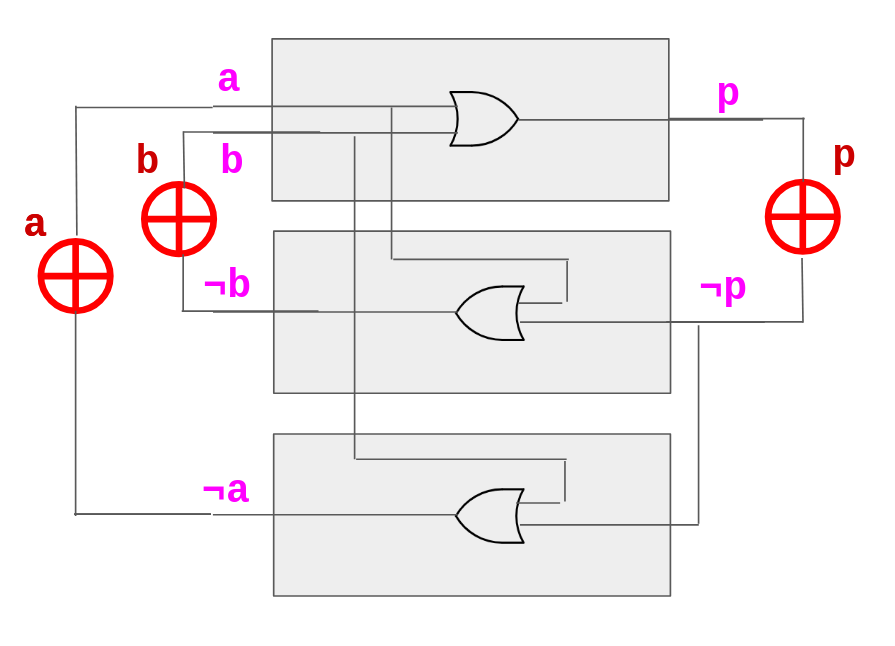}
}
\caption{Circuits and generators reproducing the semantics in standard logic of the conditional $p \rightarrow a \wedge b$.}
\label{fig:complexcircuit}
\end{figure}

Using the \(\oplus\) symbol to represent non-deterministic inputs, the overall logical system can be reproduced with the 
circuit in Fig.~\ref{fig:complexcircuit}.
This circuit exhibits interesting properties:

\begin{itemize}

\item
  the logic gates realizing the relations (the deterministic machinery)
  do not have state; in contrast, communication channels maintain state;
\item
  state transitions of the channel are determined by physical
  connections; therefore, conditionals, when used as implications, are
  not like other logic operators: they represent \textit{topological
  bindings} in the activation network;
\item
  non-deterministic inputs may determine conflicting states on the
  channel, resulting in abnormal, contradictory states of the network. In the case of logic programs, it is the ASP solver that prunes all
models involving contradictions.
\end{itemize}

\hypertarget{logical-systems-vs-logic-programs-vs-logic-gate-circuits}{%
\subsection{Logical systems vs Logic programs vs Logic gate
circuits}\label{logical-systems-vs-logic-programs-vs-logic-gate-circuits}}

Because in a logic program conditionals do not entail  their contrapositives, their semantic can be seen as more similar to that of a digital circuit compared to standard logic. Yet, there is a core difference, once again due to negation. The ``\texttt{not}'' operator in Prolog/ASP, known as \textit{negation as
failure}, enables solvers to create a conclusion out of the impossibility of deriving a conclusion. This operator is crucial to process \textit{defaults}, as it allows deriving information out of ignorance.
However, negation as failure cannot be represented as a simple operator
in a circuit, as it requires explicit machinery processing the running state of other sub-components of the system. For instance, \texttt{not p}
would correspond to a circuit whose output is true if and only if there is no active circuit activating \texttt{p}.  To summarize:

\begin{itemize}

\item
  In standard \textbf{logical systems}, strong negation is always
  present implicitly for the double negation axiom
  (\(\neg\neg a \leftrightarrow a\)); this is what enables the binding expressed by the contrapositive to emerge.
\item
  In \textbf{logic programming}, strong negation does not apply
  systematically, and so the contrapositive does not hold. Yet, negation as failure can be used in the derivation process.
\item
  In \textbf{digital circuits}, both strong and default negation are not
  defined at the system-level (there is no operationalization of the double
  negation axiom, and there is no negation as failure).
\end{itemize}
\noindent The above suggests that {we} should primarily focus on circuits relying on
AND and OR operators.

\hypertarget{activation-mechanisms}{%
\section{Activation mechanisms}\label{activation-mechanisms}}

By using only AND (conjunction) and OR (disjunction) operators in input
and in output, four types of activation patterns can be imagined. In the following, we will continue to use a Prolog-like notation, although with weaker constraints than the actual syntax. The reader is invited to interpret these rules in logic gate terms.

\hypertarget{propositional-dependencies}{%
\subsection{Propositional
dependencies}\label{propositional-dependencies}}

At first, we consider simple relations amongst simple atomic entities, corresponding in logical systems to propositional variables. The four minimal activation patterns involving three entities are: 

  \begin{enumerate}
  \def\labelenumi{(\arabic{enumi})}  
  \item
    conjunction in body: \texttt{p\ :-\ a,\ b.}
\item
    disjunction in body: \texttt{p\ :-\ a;\ b.}
  \item
    conjunction in head: \texttt{p,\ q\ :-\ a.}
  \item
    disjunction in head: \texttt{p;\ q\ :-\ a.}
  \end{enumerate}

\noindent Note that (2) is equivalent to:
\texttt{p :- a. p :- b.}
whereas (3) is equivalent to:
\texttt{p :- a. q :- a. }
The forms (2) and (3) therefore unify a number of mechanisms in a whole, whereas (1) and (4) are atomic. The form (4) is also \textit{non-deterministic} and is not allowed in
Prolog programs (but could be encoded in ASP with the choice operator,
e.g.~\texttt{1\{p; q\} :- a.}).

\hypertarget{selecting-the-most-appropriate-activation}{%
\subsubsection{Selecting the most appropriate
activation?}\label{selecting-the-most-appropriate-activation}}

Disjunction allows for both \(p\) and \(q\) to occur, therefore (4) can
in principle be seen as a more general case than (3). This redundancy
would however not hold if the disjunction is interpreted as an exclusive
disjunction (XOR). We can assume a XOR by considering additional
system-level circuitry meant to select only one alternative, as for
instance selecting the best next entity to be activated according to some metric,
e.g. most probable (e.g.~applying Bayesian probability), less complex
(e.g. according to Kolmogorov complexity), less unexpected (e.g. following
Simplicity Theory \cite{Dessalles2013}), and so on. The circuit would in this case become
deterministic, and there would be no partial overlap between (3) and (4).

\hypertarget{dependencies-between-predicates}{%
\subsection{Dependencies between
predicates}\label{dependencies-between-predicates}}

Features are always about some entity. Predicates always say something about some entity. Revising predication in terms of activation, the predicated entity would be encoded complementarily with respect to predicates (the predicate level is where the inference is operationally occurring). This separation of concerns reminds the distinction between carrier and modulating signals in signal processing; the carrier would map here to the predicated entity, the modulating signal to the predicate. This physical example suggests that, although predication traditionally refers to linguistic activities, similar arguments may be applied to describe pre-verbal, perceptual activation, as for instance in the case of mental evocation.

\hypertarget{unary-predicates}{%
\subsubsection{Unary predicates}\label{unary-predicates}}

Let us start with unary predicates. X can be seen as an object, an
event, or a situation, which is currently in focus, and being
predicated.
\begin{enumerate}
\def\labelenumi{\arabic{enumi}.}
\item
  \texttt{p(X)\ :-\ a(X),\ b(X).}
\item
  \texttt{p(X)\ :-\ a(X);\ b(X).}
\item
  \texttt{p(X),\ q(X)\ :-\ a(X).}
\item
  \texttt{p(X);\ q(X)\ :-\ a(X).}
\end{enumerate}
\noindent These dependency patterns can be illustrated by means of common examples from logic programming:
\begin{enumerate}
  \def\labelenumi{(\arabic{enumi})}
  
  \item
    is a relation that can be used to define new concepts, e.g.
    \texttt{angrydog(X)\ :-\ dog(X),\ angry(X).}
  \item
    is a relation relevant for expressing taxonomical relations, e.g.
    \texttt{mammal(X)\ :-\ dog(X);\ cat(X).} (mutual exclusion if using
    XOR)
  \item
    is a relation that activates back the source concepts from a compound
    object, e.g. \texttt{dog(X),\ angry(X)\ :-\ angrydog(X).}
  \item
    is a non-deterministic relation activating at least another concept
    (or, if using XOR, possibly a deterministic one, activating the most
    appropriate association), e.g. \texttt{dog(X);\ cat(X)\ :-\ mammal(X).}
  \end{enumerate}

Note how (1) highlights the functional presence of conceptual
\textit{merge} by morphism (e.g. take the prototype dog and make it
angrier). In logic this operation is usually operationalized as class
intersection (e.g. between the dog class and the class of angry
entities).

\hypertarget{binary-predicates}{%
\subsubsection{Binary predicates}\label{binary-predicates}}

Dependencies amongst unary predicates express possible bindings
between predicates associated with the same entity. In this paragraph, we
will discuss relationships involving multiple entities instead. In particular, We will
consider binary predicates as those used for
\textit{aggregation}, as in the proposition ``dog x has
tail y'':
$$
dog(x) \wedge tail(y) \wedge has(x, y)
$$
At the class level, these predicates are typically present in
\textit{existential rules}, as e.g.~all dogs have a tail:
$$
\forall x: dog(x) \rightarrow \exists y: tail(y) \wedge has(x, y)
$$

Let us abuse the logic programming notation by introducing existentials
with \texttt{Y/}. The four scenarios of dependency presented above
become:
\begin{enumerate}
\def\labelenumi{\arabic{enumi}.}
\item
  \texttt{p(X)\ :-\ Y/\ a(X,\ Y),\ Z/\ b(X,\ Z).}
\item
  \texttt{p(X)\ :-\ Y/\ a(X,\ Y);\ Z/\ b(X,\ Z).}
\item
  \texttt{Y/\ p(X,\ Y),\ Z/\ q(X,\ Y)\ :-\ a(X).}
\item
  \texttt{Y/\ p(X,\ Y);\ Z/\ q(X,\ Y)\ :-\ a(X).}
\end{enumerate}
Cases (1) and (2) can be treated by standard logic programming
derivation. This is because the implicit universal quantifiers are
equivalent to an existential in the body. More formally:\footnote{Note that the expression
$\forall x : [ \forall y: a(x, y) ] \rightarrow p(x) $
have a completely different meaning (e.g. if an entity has all tails, then it is
a dog).}
$$ \forall x, y: a(x, y) \rightarrow p(x) \Leftrightarrow \forall x : [ \exists y: a(x, y) ] \rightarrow p(x)
$$
In contrast, cases (3) and (4) cannot be treated with standard logic
programming derivation (nor with description logic reasoners). As before, let us interpret these mechanisms by means of examples:
  \begin{enumerate}
  \def\labelenumi{(\arabic{enumi})}
  
  \item
    is a relation determining a concept by composition, e.g.
    \texttt{car(X)\ :-\ Y/\ engine(Y),\ Z/\ wheels(Z),\ has(X,\ Y),\ has(X,\ Z).}

  \item
    can specify an operation of conceptual generalization:
    \texttt{student(X)\ :-\ Y/\ humanities(Y),\ studies(X,\ Y);\ Z/\ sciences(Z),\ studies(X,\ Z).}

  \item
    can extract parts from a whole:
    \texttt{Y/\ engine(Y),\ Z/\ wheels(Z),\ has(X,\ Y),\ has(X,\ Z)\ :-\ car(X).}

  \item
    activates possible realizations of a concept (or the most
    appropriate realization, in the case of XOR), e.g.
    \texttt{Y/\ humanities(Y),\ studies(X,\ Y);\ Z/\ sciences(Z),\ studies(X,\ Z)\ :-\ student(X).}
  \end{enumerate}
Note how (1) makes explicit the functional presence of a conceptual
\textit{merge} by composition, constructing a whole out of components.

\hypertarget{why-aggregation}{%
\subsubsection{Why aggregation?}\label{why-aggregation}}

Aggregation is an essential construct to model the world, for instance to specify attributes of classes in object-oriented programming, or, more generally, part-whole relations. Yet, looking at the literature on e.g.~mapping class diagrams to logical computational artefacts (e.g. \cite{Cali2005}), one can see how this construct requires already non-trivial machinery to be dealt with contemporary automated reasoning technologies.

Interestingly, the same construct is relevant when reasoning about causation or more generally when specifying \textit{active rules} \cite{Kowalski2016}. For instance, the following rule would be relevant to model \textit{actual causation} mechanisms (if a cause occurs, an effect follows):

\begin{lstlisting}
T2/ T2 > T1, occurs(break_window, T2) :- occurs(throw_stone, T1).
\end{lstlisting}

Remaining in the realm of causation, the form (1) is instead relevant for \textit{general causation} constructs, which would go at the meta-level with respect to actual causation, aggregating two events as cause and effect within a single causal mechanism:
\begin{lstlisting}
causal_mechanism(X) :- Y/has(X, Y), Z/has(X, Z), cause(Y), effect(Z).
\end{lstlisting}

\hypertarget{four-inferential-mechanisms}{%
\subsubsection{Four inferential
mechanisms}\label{four-inferential-mechanisms}}

Taking into account all the above, we can suggest the following
terminology for each form of dependency:




\paragraph{(1) {\textit{Comprehension}} via
\textit{merge} (as morphism, or as
composition).}

The term `comprehending' (from \textit{com-} `together' +
\textit{prehendere} `grasp') is meant to capture the idea of conceptual aggregation, which is performed by a \textit{merge} operation. Our analysis predicts two versions of merge (morphism/modification and composition/blending), in accordance with other works in cognitive studies (see e.g. \cite[p.~256]{Ghadakpour2003}). After this aggregation, comprehension activates a concept with stands for the compound.

\paragraph{(2) \textbf{\textit{Generalization}} via \textit{fusion}.}

Generalization can be seen as the inferential mechanism applied to
abstract individuals to their roles, as well as sub-roles to roles. For instance, each of us counts as researcher, any dog (or any cat, and so on) would count as an animal. In our schema, a generalization mechanism embeds alternative sub-(possibly exclusive) roles through disjunction. It relies on concepts that share a similar scaffolding. The resulting concept comes out of a stable, structural core, which can be seen as obtained by some operation akin to data \textit{fusion}. 

\paragraph{(3) \textbf{\textit{Description}} via
\textit{contrast} (as individuation, or as
instantiation).} 

Description is defined in duality to comprehension. As merge was used to
combine concepts, \textit{contrast} is used in description to disentangle
concepts. Either to extract characteristics from entities (first their
categories, and then their discriminating features); or, in the case of
compound objects, to extract their components. A perfect description would be one that allows perfect individuation/instantiation, ie. that provides a perfect match between the mental object and the features expressed by the actual object.

\paragraph{(4) \textbf{\textit{Specification} via \textit{detachment}}.}

Traditionally, inference to the best explanation is associated with
abduction. Given certain data, the observer reconstructs the underlying causes, generally applying reflective methods. However, our elaboration suggests that a similar inference can also be given a pre-verbal, perceptual interpretation. Because we are not allowing negation, we do not enter into the traditional argumentative settings (weighing pros and cons of alternative hypotheses). In non-reflective sense, we can think of the case of masked language modelling or image occlusion, i.e. inferring what would be a part of a given input which has been masked. This can be reframed as a minimal description length or minimal algorithmic information problem, and thus supports the idea of considering additional machinery for selecting the most appropriate activation. In data terms, the required operation would {be the} opposite of fusion, as it demands \textit{detaching} the ``best'' data point from a fused core.

\subparagraph{Example.}Suppose there is a murder. Evoking the concept of murder, we infer that the murderer has to be in the same place as the victim (\textit{description},  unpacking concepts that come along with ``murder''). Applying these dimensions to perceptual data (\textit{comprehension}, merging people with their spatio-temporal positions), we indicate a few people as potential suspects on the basis of these shared attributes (\textit{generalization}, extracting all common characteristics of suspects for that crime). We then select the most plausible wrongdoer (\textit{specification} --- or rather \textit{abduction}, its reflective version), singling out one instance, based on relevant deviations from that core (e.g. being the one with the clearer motive, having DNA traces on the victim, etc.).

\begin{figure}[t]
    \centering
    \scalebox{.16}{    \includegraphics{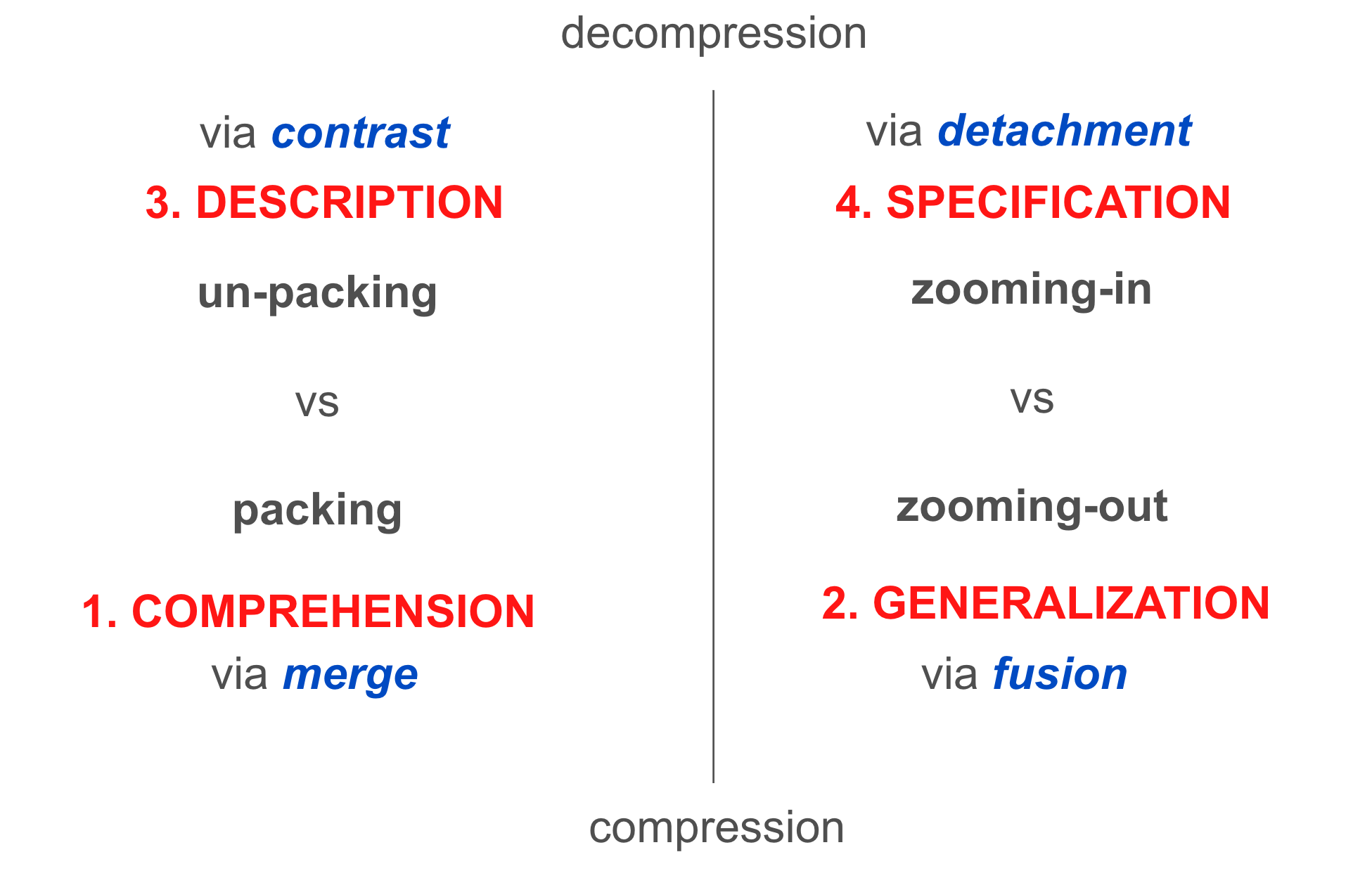}
    }
    \caption{Duality and complementarity between the inferential mechanisms 
    expressed by logic gate circuits.}
    \label{fig:schema}
\end{figure}

\subsubsection{Complementarity, 
dependency, and (de)compression}

According to the proposed analysis, comprehension (1) is dual to  description (3); generalization (2) is dual to specification (4). These two pairs are mutually irreducible, yet there may be dependencies between them. The first pair concerns \textit{packing} (by internalizing) vs \textit{un-packing} (by externalizing). The second pair can be seen in terms of \textit{zooming-out} (losing details) vs \textit{zooming-in} (gaining details). Intuitively, zooming requires having an integral space to zoom upon, therefore (2) and (4) should occur after (1). The schema expressed by the four inferential mechanisms is illustrated in Fig.~\ref{fig:schema}. The diagram can be reinterpreted in terms of information theory. Packing maps to  compression: if activated, only the output needs to be maintained, replacing all the inputs involved. This compression operation is not necessarily lossy, as it essentially corresponds to a \textit{symbolic re-encoding}. In contrast, zooming-out, for the loss of definition, maps to lossy compression, similarly to \textit{quantization}.

\hypertarget{merge-and-contrast}{%
\subsubsection{Merge and contrast}\label{merge-and-contrast}}

If the core operation for comprehension is \textit{merge}, for description would be
its inverse, \textit{contrast}. In the propositional template introduced above, the main distinction is syntactic, expressed by the presence of the conjunction (AND) on the right or the left side of the conditional:

\begin{lstlisting}
p :- a, b. % comprehension (1)
a, b :- p. % description (3)
\end{lstlisting}

Still, contrast does not play an explicit role in this expression. To make it explicit, let us approach description as an iterative, sequential process with residuals. In analogy with vectorial operations, suppose that \(a + b + c = p\), with \(a\), \(b\), and \(c\) of
decreasing size. If it can only extract one component at a time (the one
reducing the error best), description produces three sequential outputs: (i) \(a \approx p - 0\),
(ii) \(b \approx p - a\), (iii) \(c = p - a - b\). In conceptual spaces terms, $a$ may be a prototype (e.g. a dog), $b$ and $c$ modifiers (e.g. red, friendly). Or, in structural terms, $p$ may be a situational arrangement consisting of three entities, e.g. a dog $a$, a cat $b$, and a mouse $c$. As these examples show, the functioning of contrast/merge is more complex than vectorial operations (see for instance \cite{Sileno2018}). Yet, such simplification is useful for understanding the principle underlying these forms of dependency; for instance, in (1), \textit{all} activations of inputs need to be accumulated in order to trigger the output.  

\hypertarget{conjunction-and-disjunction}{%
\subsubsection{Fusion and Detachment}\label{conjunction-and-disjunction}}


As comprehension is based on merge, generalization is based on fusion, which generates a lossy intensional characterization of the inputs given to the disjunction. Specification is meant to reconstruct this input. Let us reconsider the associated dependency forms:
\begin{lstlisting}
p :- a; b. % generalization (2)
a; b :- p. % specification (4)
\end{lstlisting}
The form (2) suggests that the accumulation of activations of the common core shared between $a$ and $b$ is sufficient to trigger the output; $p$ actually acts as a placeholder for that common core. Let us suppose that $a$ and $b$ are vectors. The part they have in common can be seen as the vector which is equally distant from them, although with opposite directions, $p - a = b - p$, ie. $p = \nicefrac{a + b}{2}$. In order to be able to fully reconstruct the original points however, we also need to add a range, some spatial information around the center, e.g. $q = \nicefrac{|a - b|}{2}$. The entity $p$ resulting from generalization would have both a center and some regional information. Detachment can then work the other way around: we start from an activated generalized entity (e.g. a prototype), defined by a center (a default) and some range (expected deviations), and we have only to select in which direction to move. In complete ignorance, any direction would be equally fine. To reduce prediction errors in unmasking, models/methods presented in the literature consider the probability of occurrence, or conceptual accessibility, or other weighting mechanisms to be in place. Note that this prioritization would map at a meta-level with respect to the structural activation mechanisms discussed here. Rather than the topology of the circuit, this would concern the intensity of tensions/current propagated in the circuit.

\hypertarget{learning}{%
\subsection{Learning}\label{learning}}

Inference is a cognitive process allowing agents to interpret perceptual data, possibly fill in gaps, and make sense of the world despite uncertainty or missing information. In our simplistic model, the inferential systems consist of circuits corresponding to rule-like structures. Sensory inputs and the topology of concepts are fixed at inference time. Learning, as a process modifying the inferential system, would map to adding or removing a rule in one of the four template forms, or rather, in circuit terms, to add or remove the topological connection expressed by the implication of that rule. 


From an informational point of view, new concepts are introduced either as a re-encoding of merge or as a fusion of sensory inputs and/or of other concepts. Learning concerns, therefore, comprehension and generalization. Comprehension (1) has primarily an extensional nature: a number of entities have to be activated for the output to be generated. Generalization (2) has instead an intensional nature, as the generalizing structure to be captured for triggering activation is only implicitly defined by the input items. An additional distinction between the two can be made by considering the associationist principle that supports the emergence of a new concept. For (1), the triggering of the output is determined by the co-occurrent activation of the inputs; therefore, it builds upon \textit{positive} associations. The more often two entities are jointly activated, the more it makes sense to introduce a compound entity. For (2), the triggering of the output is determined by the common structural core. Therefore, the more often this common structural core occurs, the more it makes sense to introduce the generalized entity. But which entities should we attempt to combine when they typically do not occur at the same time? Interestingly, the exclusive reading of the disjunction (XOR) suggests capturing also \textit{negative} associations between the inputs, offering a heuristic to bootstrap this selection phase. 

\subparagraph{Example} Suppose a dataset is given with a finite number of numeric features. The dataset in itself can be seen as resulting from reifying observations (\textit{description}), one for each instance of the dataset. We then consider these dimensions (\textit{comprehension}) to create a vectorial space. General information of the dataset can be obtained by computing the mean and stdev for each dimension (\textit{generalization}).
Yet, to gain more knowledge about the dataset, we may apply a clustering algorithm. Clustering algorithms build upon both positive and negative associations. They generally aim for low \textit{intra-cluster} distance (points within a cluster are similar, ie. they have much in common -- these features would be inputs for a \textit{comprehension} mechanism) as well as for a high inter-cluster distance (clusters are well separated, ideally covering mutually exclusive regions -- these would be input for a \textit{generalization} mechanism). The trade-off heuristics balancing these two layers will determine the actual clusters.

\hypertarget{probabilistic-interpretation}{%
\section{Probabilistic
interpretation}\label{probabilistic-interpretation}}

So far, we have considered digital circuits, specified as Prolog-like rules, where inputs and outputs can only have Boolean states. In this section, we briefly elaborate on an extension informed by the semantics of probabilistic programs. 

\hypertarget{probabilistic-programs}{%
\subsection{Probabilistic programs}\label{probabilistic-programs}}

In
ProbLog \cite{DeRaedt07}, Prolog-like rules are given probabilistic information:
\begin{lstlisting}
0.3 :: b :- a.
\end{lstlisting}
This rule can be interpreted as $P(b|a) = 0.3$.\footnote{Rewriting a rule with its probability seems to be aligned with Adams's thesis, ie. that the acceptability of a rule is given by its conditional probability; yet, this also brings also all the critiques made in philosophy towards this assumption, see e.g. \cite{Hajek2012}.} The same applies for facts. For instance,
\begin{lstlisting}
0.7 :: c.
\end{lstlisting}
it has to be interpreted as $P(c) = 0.7$. With a probabilistic interpretation, all propositions become truthbearers with a certain degree. Following the duality holding in probability and in logic (due to the underlying extensional semantics), we have: $P(\neg c) = 1 - P(c) = 1 - 0.7 = 0.3$. 

Integrating probability theory with symbolic rules enables us to define the inferential system as a system of continuous functions (bringing relevant properties like differentiability). This possibility has recently attracted a renewed interest in neuro-symbolic approaches, for instance to add explicit safety constraints to reinforcement learning \cite{Yang2023}.  

In our simplistic electronic interpretation, such an extension would allow us to pass from a digital system (only 0 or 1 electric states) to an \textit{analogical} system: the varying values of tension would be aligned to the associated probability value.

\hypertarget{dependencies-amongst-dependencies}{%
\subsection{Dependencies amongst
dependencies}\label{dependencies-amongst-dependencies}}

We can then specify the four types of dependency in probabilistic terms by
applying set operations. The resulting probability values would be proportional to tensions propagated by the circuit.

\begin{enumerate}
\def\labelenumi{\arabic{enumi}.}

\item
  \(P(p|a \wedge b)\)
\item
  \(P(p|a \vee b) = P(p|a) + P(p|b) - P(p|a \wedge b)\)
\item
  \(P(p\wedge q|a) = P(p|q \wedge a) \cdot P(q|a) = P(q|p \wedge a) \cdot P(p|a)\)
\item
  \(P(p\vee q|a) = P(p|a) + P(q|a) - P(p \wedge q|a)\)
\end{enumerate}

\noindent We could also consider the cases with XOR instead of OR, knowing that:
\(p \oplus q = (p \wedge \neg q) \vee (\neg p \wedge q)\):
\begin{enumerate}
\def\labelenumi{\arabic{enumi}.}
\setcounter{enumi}{4}

\item
  $P(p | a \oplus b) = P(p | (a \wedge \neg b) \vee (\neg a
  \wedge b)) = P(p |  a \wedge \neg b) + P(p | {}\neg a
  \wedge b) $
\item
  $P(p \oplus q |  a) = P((p \wedge \neg q) \vee (\neg p
  \wedge q) |  a) = P(p \wedge \neg q |  a) + P(\neg p
  \wedge q |  a) $
\end{enumerate}

\noindent From the expressions above, we see that:
\begin{itemize}
\item
  (4) and (6) depend on (3)
\item
  (3) depends on (1)
\item
  (2)  and (5) depend on (1)
\item
  (1) is independent of all other forms.
\end{itemize}
Assuming this numeric model to be cognitively informative, we may build an inferential system handling all these mechanisms. To do so, we should first operationalize  \textit{comprehension} (with \textit{merge}), then
\textit{generalization} (with \textit{fusion}) and in parallel \textit{description}, and from these
ingredients finally perform \textit{specification}. This elaboration suggests that abduction (as a reflective form of specification) is the highest possible level of inference, whereas lower forms of induction can already be introduced with comprehension, and subsequently with generalization.

\subsection{Discriminative vs Generative}

Machine learning models are traditionally distinguished between: (i) discriminative models (e.g. given an image, the algorithm qualifies it with a label/class, on the basis of a certain \textit{heuristics}); (ii) generative models (e.g. given a class, the algorithm generates an image, on the basis of certain \textit{constructors}). Heuristics and constructors are associated with different informational principles. If $y$ is the class, and $x$ is the object, discriminative models are built estimating $P(y|x)$, generative models are built estimating $P(x, y)$. To control generative models, we set the $y$, and then $x$ can be extracted following $P(x|y)$.

It is easy to see that the generative case maps to \textit{specification} (4)  (e.g. text completion is a form of unmasking) rather than \textit{description} (3), which may hint to an architectural flaw. In contrast, the discriminative case maps intuitively to \textit{generalization} (2) (what counts as a certain class is intensionally determined by the data points in that class). However, for a non-binary classification task, the final inferential mechanism can also be seen as \textit{specification} (4), because it should select the class, amongst the available ones, whose activation triggered by the input is stronger. These examples show that more complex inferential tasks consist of several components. Further study is needed to explore the various patterns that emerge across different AI methods and cognitive models. 


\section{Perspectives}

The early days of AI started by attempting to connect artificial neural networks with digital circuits. The McCulloch-Pitts neuron \cite{mcculloch1943logical}, besides paving the way to the Perceptron and eventually to contemporary machine learning, inspired \textit{threshold logic}, which later found its way back in electronics. Obviously, this model is highly limited for capturing the complexity of contemporary AI architectures, even more the neural activity of actual brains. 

Yet, the simplicity of the mapping between logic ports and inferential mechanisms still facilitates speculation and imagination. The scattered view that we may have of cognitive mechanisms, both natural and artificial, makes the demand for unifying theories still a valid endeavour. If unifying theories fail with simplistic models, they would fail even more with realistic models. This principle explains why our elaboration took an opposite stance compared to contemporary approaches, which try to make sense of what machines do at the artificial neuronal level, or what human brains do at an electro-physiological level. 

In this paper, we assume that reproducing higher-level functions of cognition with symbolic AI in material form is relevant to represent functions realized by the mind. In reference to the \textit{structural/functional} distinction in cognitive systems discussed in \cite{Lieto2021}, we do adhere to the functional realm (we do not make any reference to the biological counterpart), though we keep a structural view of the informational system. Under this assumption, we were able to construct a more unifying picture, predicting the presence of higher-level operations like merge, contrast, fusion, and detachment, in support of four mechanisms: comprehension, description, generalization, and specification. We were able to provide an analysis of mutual dependencies, hypothesizing an order in which these functions emerge in inferential systems. 

What precedes is little more than a sketch, though, the first elaborations of a new conjecture. Further experiments are needed to consolidate these results (e.g. running systems inspired by this decomposition, additional examples from existing methods in symbolic and sub-symbolic AI and established cognitive models). A parallel stream of work concerns going beyond probabilistic methods for the proposed analogical extension, such as using methods informed by algorithmic information theory \cite{sileno2021,sileno2023}. 

\bibliography{references}

\end{document}